\pdfoutput=1
\documentclass[11pt]{article}

\usepackage[a4paper,margin=1in]{geometry}
\usepackage{newtxtext,newtxmath}
\usepackage{graphicx}
\usepackage{booktabs}
\usepackage{amsmath}
\usepackage{url}
\usepackage{enumitem}
\usepackage{caption}
\usepackage{float}
\usepackage{placeins}
\usepackage{subcaption}
\usepackage{algorithm}
\usepackage{algpseudocode}
\usepackage{xurl}
\usepackage[hidelinks]{hyperref}
\usepackage[numbers]{natbib}

\title{
Runtime Evaluation of Procedural Content Generation in an Endless Runner Game Using Autonomous Agents
}

\author{
Rishabh Kar\\
Department of Informatics\\
King's College London\\
London, United Kingdom\\
\texttt{rishabh.kar@kcl.ac.uk}
}

\date{}

\begin{document}

\maketitle

\begin{abstract}
Procedural Content Generation (PCG, the practice of creating game content such as terrain, levels, and objects through algorithmic rules rather than by hand) enables game content to be created algorithmically without direct manual level-design effort, but it introduces a serious evaluation problem: generated content may become unbalanced, blocked, repetitive, or technically unsolvable. This paper presents \textit{Momentum}, an endless-runner game that integrates runtime terrain generation, environment object spawning, and autonomous agent-based evaluation into a single gameplay loop. Ground tiles and environmental objects are generated dynamically as the player advances, object placement follows a constraint-driven mechanism inspired by Wave Function Collapse (WFC, an algorithm that fills a grid by repeatedly choosing a value for each cell while respecting compatibility rules with its neighbours), and the runtime navigation surface is rebuilt asynchronously to remain consistent with the streamed environment. Two autonomous evaluation agents move ahead of the player and inspect the generated path: an aerial scanner that examines the corridor geometrically, and a ground-traversal agent that validates the same region from a navigational perspective. The evaluation pipeline combines ray casting (firing a virtual line into the scene to detect what it hits), volumetric physics sweeps (testing whether a 3D box-shaped region overlaps any solid object), obstacle-layer filtering (restricting detection to objects tagged as obstacles), and structured crash reporting to identify problematic generated scenarios before they reach the player. The work demonstrates how generation and validation can be unified within the same runtime loop, rather than treating evaluation as a separate offline pass. Around this implementation, the paper formulates a measurable evaluation framework along the canonical PCG axes of playability, diversity, controllability, and runtime performance, derives a structural saturation bound on the spawner from its own placement constraints, and quantifies the per-segment scanning cost of the agents from first principles.
\end{abstract}


\section{Introduction}

\subsection{Motivation}

Procedural Content Generation offers a practical means of producing game content through algorithmic rules rather than complete manual construction. In game development, this reduces the effort required to author levels, terrain, objects, and environmental variation, particularly in genres where replayability and unpredictability are central to the experience. Endless-runner games present a particularly demanding case for this approach because their content requirement is continuous: the player advances indefinitely, and the system must continuously produce a playable path. A purely handcrafted level rapidly becomes repetitive, while a generated one introduces variation at the cost of admitting blocked paths, unsuitable object placements, and sections that cannot be completed.

The motivation for this work is therefore not generation alone, but the integration of generation with evaluation. A procedurally generated section may appear visually acceptable while remaining technically invalid for the player. \textit{Momentum} addresses this through autonomous agents that inspect the generated path before the player reaches it, treating evaluation as a first-class component of the runtime rather than a post-hoc check.

\subsection{Problem Statement}

The central problem addressed in this work is the reliability of procedurally generated content in a runtime environment. Procedural generation does not, in itself, guarantee that the produced content is playable. A generation algorithm may place objects in a manner that obstructs the route, interrupts navigation, or yields a section that the player cannot reasonably traverse. This produces a technical gap between content creation and content validity: producing terrain and objects is not sufficient if the system cannot determine whether those generated sections remain safe for traversal.

The problem becomes more acute in a runtime setting, where content is produced while the player is already moving through the world and no offline authoring pass is available. The problem can therefore be stated in two coupled parts. First, the system must generate terrain and environmental objects dynamically so that gameplay can continue beyond fixed level boundaries. Second, the same system must evaluate whether the generated path remains technically sound before the player encounters it. This paper studies that problem through \textit{Momentum}, in which autonomous aerial and ground agents inspect the generated path during gameplay.

\subsection{Contributions}

This paper makes the following contributions:

\begin{enumerate}[leftmargin=*]
    \item It presents \textit{Momentum}, an endless-runner game in which the level is generated during runtime rather than being constructed through manual level design.
    \item It implements a procedural generation pipeline that streams ground tiles and environment objects dynamically as the player advances, with asynchronous rebuilding of the navigation surface to maintain consistency with the moving world.
    \item It applies a constraint-driven object placement method, inspired by Wave Function Collapse, that distributes objects across the generated ground while preserving an intended traversable lane.
    \item It integrates runtime gameplay controls for modifying player speed, lateral movement speed, and object spawn density, allowing the behaviour of the generator to be perturbed during execution.
    \item It introduces autonomous evaluation agents that move ahead of the player and inspect the generated path before the player reaches that section of the level.
    \item It combines ray casting, volumetric physics checks, and obstacle-layer filtering to identify blocked paths, unsafe object placements, and technically invalid generated sections.
    \item It implements a crash-reporting pipeline that records blockage details, player state, generation parameters, and offending objects, and presents this information in a report for later analysis.
\end{enumerate}

\subsection{Paper Organisation}

The remainder of this paper is organised as follows. Section 2 presents the background and related work on Procedural Content Generation, game content evaluation and the techniques relevant to this project. Section 3 defines the objectives and technical specification used to guide the development of the system. Section 4 describes the design, methodology and implementation of \textit{Momentum}, including the system architecture, player physics, terrain generation, object spawning, skybox variation and runtime interface.

Section 5 presents the analysis and evaluation mechanism, focusing on the autonomous aerial agent, ground agent, ray casting, volumetric checks, blockage detection and crash-reporting pipeline. Section 6 discusses the legal, social and ethical considerations associated with procedural generation and automated evaluation in games. Section 7 concludes the paper.

\section{Background and Related Work}

\subsection{Procedural Content Generation}

Procedural Content Generation (PCG) is a design approach that uses algorithmic processes to create game content. It was initially adopted in early titles such as \textit{Rogue} and \textit{Maze Craze}, where limited system resources were the primary constraint. In that context, PCG offered an efficient solution by using seeds and rules to generate larger worlds dynamically while reducing memory and storage requirements.

PCG has since moved beyond a workaround for hardware limitations. As computing power increased, developers adopted procedural methods to automate parts of the content creation workflow and to expand the variety of content produced. Procedural techniques are now used to generate textures, three-dimensional objects, level layouts, terrain, vegetation, weather patterns, and other environmental detail, with applications extending to simulation, film production, architectural visualisation, virtual reality, and extended reality. A compact rule set or seed can produce expansive worlds without requiring every asset to be stored or every level to be authored manually, which improves scalability and increases unpredictability. Titles such as \textit{Minecraft}, \textit{No Man's Sky}~\cite{mcaloon2016nomanssky}, and \textit{Caves of Qud}~\cite{bucklew2017qud, bucklew2019qud} are widely cited examples of how PCG can produce large worlds in which the experience remains fresh across sessions.

The drawbacks of PCG are equally well-known. Procedurally generated environments can become monotonous when framed by simple rules, and they often lack the handcrafted detail that human designers introduce during manual level design. More importantly, automatically generated content can become unsolvable, poorly balanced, or technically invalid. Debugging such systems is difficult because randomness obscures the conditions that produced a given failure, and reproducing a specific defect requires reproducing the exact seed and parameter set that caused it. PCG should therefore not be treated solely as a content-generation method; it requires evaluation and validation techniques to determine whether the generated content is technically sound. In this work, PCG is used to generate the game environment during runtime, while the evaluation mechanism examines whether the generated path remains playable before it affects the player.

\subsection{Procedural Terrain Generation Techniques}

There are several ways in which procedural terrain can be generated, and the choice usually depends on what the game is trying to create. One of the foundational approaches towards terrain generation is the use of noise functions to create fractal height maps. Earlier work used the diamond-square algorithm for midpoint displacement, and Perlin Noise and its successor forms are still widely used for creating synthetic terrains with multiple variations.

Noise-based generation works well when the terrain needs rough surfaces, height variation and natural-looking topology. By combining different frequencies and amplitudes, the system can create rough height maps that imitate real-world terrain patterns. These techniques are also memory efficient and can be controlled through different parameters. However, a major disadvantage appears when uniform noise starts creating repeated patterns. Designers often need to combine noise signals with other effects to make the output look more natural. This project does not use noise-based terrain generation because the terrain is made of constant ground tiles, where height variation is minimal. Noise is more suitable for content that is topographically rich.

Grammar-based generation is another approach used for procedural content. Generative grammars create content by applying rewriting rules, similar to the grammar mechanism used to construct sentences. This allows developers to focus on structural patterns rather than placing every object manually. A common example is L-Systems, which use recursive rules to grow plants and branching structures. Graph grammars and shape grammars have also been explored for generating mission paths and spaces~\cite{dormans2010level}.

The advantage of grammar-based generation is that it allows designers to encode structure and validity through rules. However, there is still a chance of repetitiveness if the same rules are applied repeatedly~\cite{samaritaki2022grammars}. It is more suitable for environments that contain complex objects, vegetation or structured spaces. In this project, grammar-based generation was considered as part of the literature, but it was not required for the current lane-based endless-runner environment.

Wave Function Collapse is comparatively newer than the other alternatives and has gained popularity because of its ability to generate new data from examples~\cite{buyuksar2023wfc}. The technique works by keeping several possible states for each cell and then gradually reducing those possibilities while maintaining compatibility with neighbouring cells. In this project, Wave Function Collapse is not used as a complete terrain generation system. Instead, it is used as an inspiration for placing environmental objects on the generated ground tiles, so that the placement is controlled but still not fully predictable.

\subsection{Wave Function Collapse}

Wave Function Collapse (WFC) is one of the procedural generation methods reviewed for this project. The method is notable for its ability to generate new content from a small set of examples while preserving local compatibility between neighbouring elements. It was first proposed by Merrell in 2007 and later popularised through Gumin's implementation in 2016.

At a conceptual level, WFC operates on a grid. Each cell begins with a set of possible states, and these possibilities are reduced as constraints from neighbouring cells are propagated. The final output is therefore not purely random: it retains variation, but each local choice must remain compatible with the surrounding structure. Karth and Smith~\cite{karth2017wfc} characterise WFC as a form of constraint solving in which local compatibility rules can produce coherent global patterns. This interpretation is relevant to the present work, since the aim is not unconstrained randomness but object placement that varies while still respecting simple spatial constraints.

In \textit{Momentum}, WFC is not used as a complete terrain-generation method, nor is it implemented through a neural network or trained machine learning model. The implementation is a simpler one-dimensional, WFC-inspired object placement mechanism. For each generated ground tile, the lateral axis is divided into grid cells. Cells are selected for object placement, neighbouring cells are marked as occupied to reduce clustering, and the traversable lane is preserved through a dedicated clearance parameter. A ray cast then confirms that valid ground exists before an object is instantiated. Object spawning is handled by a dedicated spawn manager that controls the lifecycle of obstacle prefabs already present in the project assets. These prefabs are grouped according to skybox variants, allowing the spawned objects to remain visually consistent with the active environment theme. WFC is therefore used as a controlled object-placement strategy rather than as a complete map-generation algorithm.

\subsection{Evaluation Mechanisms for Procedural Content}

Evaluation is required because procedural generation does not automatically guarantee that the generated content is valid or playable. In games, runtime validation is used to check whether generated content can actually be used by the player, often by simulating some form of player interaction~\cite{togelius2008automatic}. The basic requirement in this type of scenario is reachability, because the generated section should not block the player from moving forward or continuing the game.

There are different ways in which procedural content can be evaluated. Static analysis checks the validity of a level before runtime, while automated agents can perform walk-through mechanisms either before or during runtime to identify problematic areas~\cite{cook2024evaluation}. Another approach is an adaptive feedback loop, where the system reacts to the generated content during gameplay and adjusts the generation process based on that feedback~\cite{cook2022danesh}.

For this project, the focus is on runtime evaluation using automated agents. The reason for this is that the game content is created dynamically while the player is moving, so the evaluation also needs to happen during execution. The agents move ahead of the player and check whether the generated path remains technically playable. This makes the evaluation mechanism part of the game loop rather than a separate post-processing step.

The evaluation also uses geometric scanning methods such as ray casting and volumetric checks. Ray casting is useful for checking gaps, obstacle distribution and ground availability by projecting rays into the scene. Volumetric checks are used along with ray casting to improve detection of objects that may block the path. Together, these methods help identify blocked sections, unwanted object placement and unsolvable generated scenarios before they become a direct problem for the player.

\section{Objectives and Technical Specification}
\label{sec:objectives}

The primary objective of this work is to design and implement a content generation system that operates within the runtime of a game, together with a mechanism that evaluates the generated content before it reaches the player. The procedural generator is expected to produce levels, ground sections, and game objects automatically, without depending on manual level-design effort. The evaluation mechanism is expected to examine the generated path during runtime and report sections that are blocked, unsolvable, or otherwise technically invalid. The two objectives are coupled by design: a generator that emits invalid content is of limited value, and an evaluator that operates only on completed levels does not apply when content is produced incrementally during play.

The system must satisfy six concrete requirements: (i) continuous streaming of terrain and objects ahead of the player; (ii) randomisation of object placement sufficient to preserve replayability while reserving a viable traversable lane; (iii) deployment of autonomous aerial and ground agents that inspects upcoming sections before the player reaches them; (iv) the option to rectify detected blockages so that an isolated invalid placement does not terminate the run; (v) live runtime controls for player speed, lateral speed, and spawn density; and (vi) structured recording of any detected failure together with the parameters that produced it. Stability of frame rate, modular code organisation, and adherence to engine-specific style conventions~\cite{unity2025style} are treated as engineering requirements rather than research contributions.

\begin{table}[H]
\centering
\caption{Technical specification of the project.}
\label{tab:technical_specification}
\begin{tabular}{ll}
\toprule
Component & Specification \\
\midrule
Game engine & Unity 6000.2.0b6 \\
Programming language & C\# \\
Game type & 3D endless-runner game \\
Generation system & Runtime terrain and object generation \\
Object placement & Wave Function Collapse-inspired spawning \\
Navigation system & Unity NavMesh with asynchronous baking \\
Evaluation mechanism & Autonomous agents, ray casting, and volumetric checks \\
Physics support & Engine physics and collider-based detection \\
User interface & Runtime sliders and in-game feedback display \\
Reporting system & Crash-report display and PDF export \\
\bottomrule
\end{tabular}
\end{table}
\FloatBarrier

\subsection{Evaluation Questions, Hypotheses, and Metrics}
\label{sec:eval-questions}

The objectives in Section~3 specify what the system should do; they do not specify how its behaviour should be measured. Following the taxonomy of Cook, Withington, and Tokarchuk~\cite{cook2024evaluation}, who organise PCG evaluation around \emph{quality/playability}, \emph{diversity}, \emph{controllability}, and \emph{performance}, this work formulates four research questions and pairs each with an explicit hypothesis and a measurable quantity that is recoverable either directly from the code or from runtime instrumentation.

\begin{description}
  \item[\textbf{RQ1 (Playability).}] Does agent-based runtime evaluation reduce the rate at which unhandled blocked sections reach the player, relative to a generator running without evaluation? \emph{Hypothesis~H1:} the auto-removing aerial scanner is expected to reduce the number of player-encountered blockages, while diagnostic-only configurations leave the underlying blockage rate unchanged but make it visible.
  \item[\textbf{RQ2 (Controllability \& Diversity).}] Does the spawn-density parameter $p_{\mathrm{spawn}}$ produce monotonic, predictable changes in placement count and observable changes in prefab diversity? \emph{Hypothesis~H2:} the realised spawn count saturates well below the value requested by Equation~(25) once the lane-clearance and adjacency constraints bind, so the slider exhibits a controllability ceiling that is structural rather than parametric.
  \item[\textbf{RQ3 (Performance).}] Can both agents perform their per-tile inspection within the $16.66~\mathrm{ms}$ frame budget associated with $60~\mathrm{FPS}$, or at least within the $33.33~\mathrm{ms}$ budget associated with $30~\mathrm{FPS}$, as defined by Unity's profiling guidance~\cite{unityProfileBestPractices,unityProfileBlog}? \emph{Hypothesis~H3:} the per-segment cost is bounded by a small constant independent of run length, because the scanner is gated by an integer tile index and amortises the OverlapBox sweep across many ray probes.
  \item[\textbf{RQ4 (Coverage complementarity).}] Do the aerial and ground agents detect overlapping or complementary failure sets? \emph{Hypothesis~H4:} the two agents inspect distinct properties of the same region (geometric clearance vs.\ NavMesh traversability) and therefore each detects failures the other does not, so the union of their reports is strictly larger than either taken alone.
\end{description}

The metrics used to address these questions are the following. Let $S_{\mathrm{total}}$ be the total number of corridor scan segments inspected during a run and $S_{\mathrm{blocked}}$ those classified as blocked. The blockage and passability rates are
\begin{equation}
  R_{\mathrm{block}} = \frac{S_{\mathrm{blocked}}}{S_{\mathrm{total}}},\qquad
  R_{\mathrm{pass}} = 1 - R_{\mathrm{block}}.
  \label{eq:blockage}
\end{equation}
For runs with auto-removal enabled, with $B_{\mathrm{detected}}$ the set of detected blockers and $B_{\mathrm{removed}}\subseteq B_{\mathrm{detected}}$ those destroyed by the scanner before the player arrived, the auto-removal rate is
\begin{equation}
  R_{\mathrm{remove}} = \frac{|B_{\mathrm{removed}}|}{|B_{\mathrm{detected}}|}.
  \label{eq:remove}
\end{equation}
Ground-agent recoveries are normalised by distance travelled,
\begin{equation}
  R_{\mathrm{recover}} = \frac{N_{\mathrm{rec}}}{D}\quad[\mathrm{recoveries\,m^{-1}}].
  \label{eq:recover}
\end{equation}
The spawn realisation ratio compares actually-instantiated objects $O_{\mathrm{spawned}}$ against the count requested by Equation~(18),
\begin{equation}
  \rho_{\mathrm{spawn}} = \frac{O_{\mathrm{spawned}}}{O_{\mathrm{intended}}},\qquad
  O_{\mathrm{intended}} = \left\lfloor N \cdot \tfrac{p_{\mathrm{spawn}}}{100} \right\rfloor,
  \label{eq:realisation}
\end{equation}
with $N$ the cell count from Section~4.5. The frame-budget violation rate~\cite{unityProfileBestPractices} is
\begin{equation}
  V(\tau) = \frac{1}{F}\sum_{i=1}^{F}\mathbb{1}[f_i > \tau],\quad \tau\in\{16.66,\,33.33\}~\mathrm{ms}.
  \label{eq:violation}
\end{equation}
Prefab diversity within a run is captured by the Shannon entropy of the empirical prefab distribution and its normalised form~\cite{cook2024evaluation},
\begin{equation}
  H_{\mathrm{prefab}} = -\sum_{k=1}^{K} p_k\log_2 p_k,\qquad
  H^{\mathrm{norm}}_{\mathrm{prefab}} = \frac{H_{\mathrm{prefab}}}{\log_2 K}\in[0,1].
  \label{eq:entropy}
\end{equation}
Agent-set complementarity is measured by the Jaccard overlap of their reported blocker sets $A$ and $G$,
\begin{equation}
  J(A,G) = \frac{|A\cap G|}{|A\cup G|},\qquad
  C_{\mathrm{unique}} = 1 - J(A,G),
  \label{eq:jaccard}
\end{equation}
where smaller $J$ (equivalently, larger $C_{\mathrm{unique}}$) indicates greater complementarity between the two evaluators.

Table~\ref{tab:metric-mapping} maps each research question to the metric and to the implementation source from which the measurement is obtained.

\begin{table}[H]
\centering
\caption{Mapping between research question, metric, and implementation source.}
\label{tab:metric-mapping}
\footnotesize
\begin{tabular}{p{0.55cm} p{3.9cm} p{4.3cm} p{5.5cm}}
\hline
\textbf{RQ} & \textbf{Metric} & \textbf{Measurement} & \textbf{Source in implementation} \\
\hline
RQ1 & $R_{\mathrm{block}}$, $R_{\mathrm{remove}}$, player-encountered blockage count & Counters over scan segments and reported blockers per run & \texttt{FlyerCorridorScanner.ScanCorridor},\newline \texttt{BlockageReporter.ReportBlockage},\newline auto-removal branch \\[4pt]
RQ2 & $\rho_{\mathrm{spawn}}$, $H^{\mathrm{norm}}_{\mathrm{prefab}}$, $R_{\mathrm{block}}$ vs.\ $p_{\mathrm{spawn}}$ & Per-tile spawn counters; histogram over instantiated prefab IDs & \texttt{EnvironmentObjectSpawnManager.}\newline\texttt{SpawnObjectsOnGround} \\[4pt]
RQ3 & Mean / p95 / p99 frame time, $V(16.66)$, $V(33.33)$ & Per-frame $\Delta t$ samples over a fixed-length run & Unity frame loop / Profiler API \\[4pt]
RQ4 & $J(A,G)$, $C_{\mathrm{unique}}$, $R_{\mathrm{recover}}$ & Comparison of reported blocker sets between the aerial and ground agents & \texttt{FlyerCorridorScanner} vs.\newline \texttt{GhostRunnerAgent.ScanCorridor} \\
\hline
\end{tabular}
\end{table}
\FloatBarrier

\section{Design, Methodology, and Implementation}

\subsection{System Architecture}

The system architecture of \textit{Momentum} is modular, so that gameplay logic, procedural generation, and evaluation are not coupled within a single script. The project is decomposed according to behaviour: each component is responsible for a clearly defined part of the system, including player movement, camera handling, terrain generation, environment object spawning, skybox variation, runtime UI controls, autonomous agents, navigation-mesh management, blockage reporting, and crash-report export.

The player controller manages forward motion, lateral movement, jumping, and interaction with the engine's physics system. The terrain generation component spawns ground tiles ahead of the player, and the object spawning component places environmental objects on the generated ground. The skybox manager coordinates with the generation pipeline by varying the environment theme as the player progresses, and the object spawner consults the active theme when selecting prefabs.

The evaluation side of the architecture is implemented through autonomous agents. The aerial agent moves ahead of the player and scans the generated corridor from above, while the ground agent validates the same region through the runtime navigation surface. Both agents are supported by ray casting, volumetric physics checks, and obstacle-layer filtering. When a blocked or unsolvable section is detected, the relevant data is forwarded to the blockage reporting subsystem.

The reporting pipeline stores the detected information in a structured crash-report record and carries it across the scene transition into the exit-report scene. The report is rendered within the game and may also be exported as a PDF document. The architecture therefore not only generates content, but also provides a feedback mechanism for analysing where and why the procedural generation process produced an invalid section.

\begin{figure}[t]
    \centering
    \includegraphics[width=0.95\linewidth]{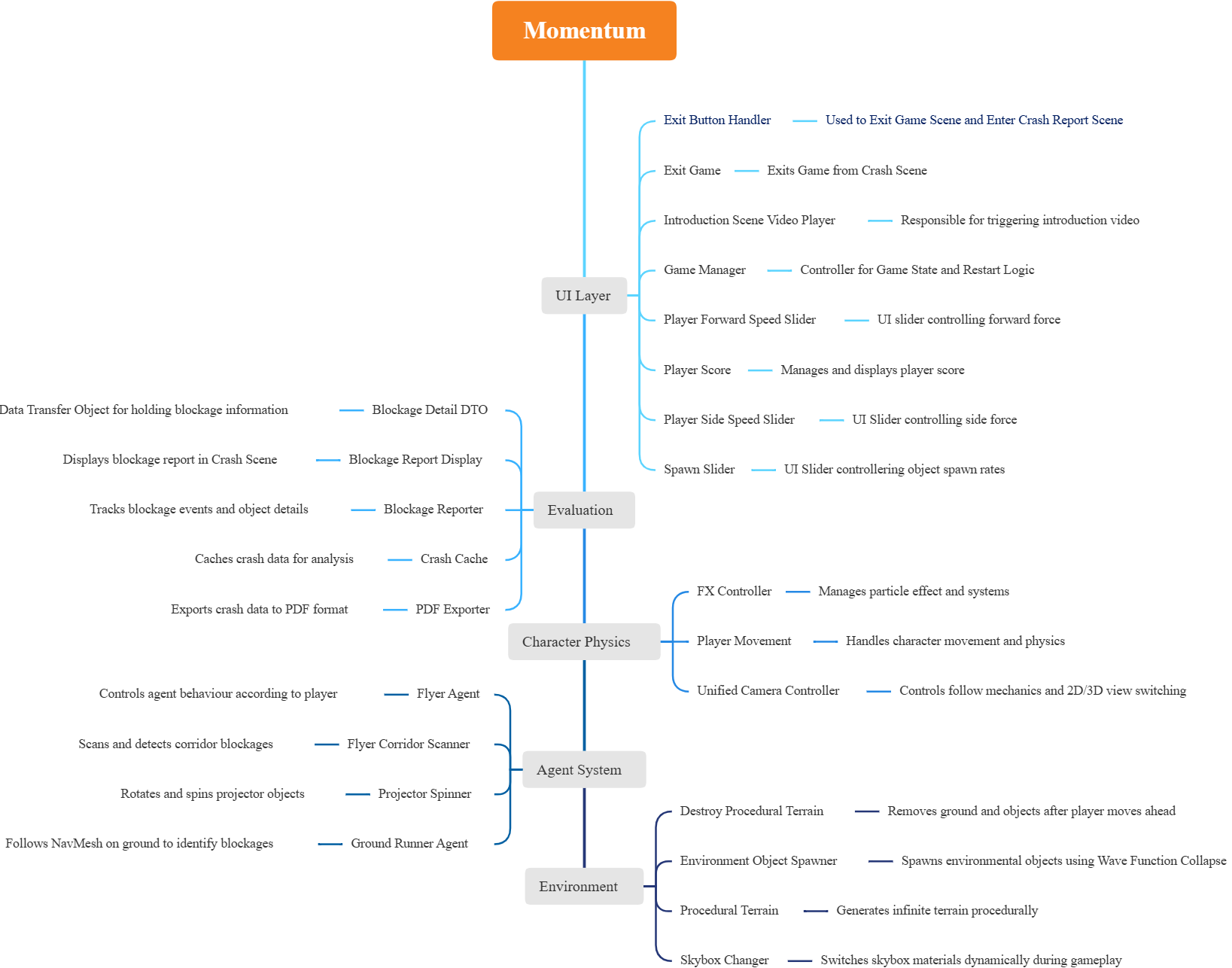}
    \caption{High-level architecture of the \textit{Momentum} system, organised into the UI layer, evaluation layer, character physics, agent system, and environment.}
    \label{fig:architecture}
\end{figure}

\subsection{Game Concept}

The game developed in this project is called \textit{Momentum}. It is implemented as an endless-runner in which the player advances continuously and the surrounding environment is generated during gameplay. The player is not presented with a fixed level to complete; instead, the run continues until the player fails or exits, and the score increases with forward progression.

The concept is intentionally minimal so that the focus of the work remains on procedural generation and runtime evaluation. The player can move laterally and jump, while the ground and objects are generated ahead of the player. This makes the game a suitable test bed for dynamic generation, since the system is not dependent on a pre-authored level. Endless-runner games are particularly well suited to this purpose because their content requirement is continuous: a repeated handcrafted level becomes predictable within minutes, whereas runtime generation can sustain variation indefinitely.

\begin{figure}[t]
    \centering
    \includegraphics[width=1\linewidth]{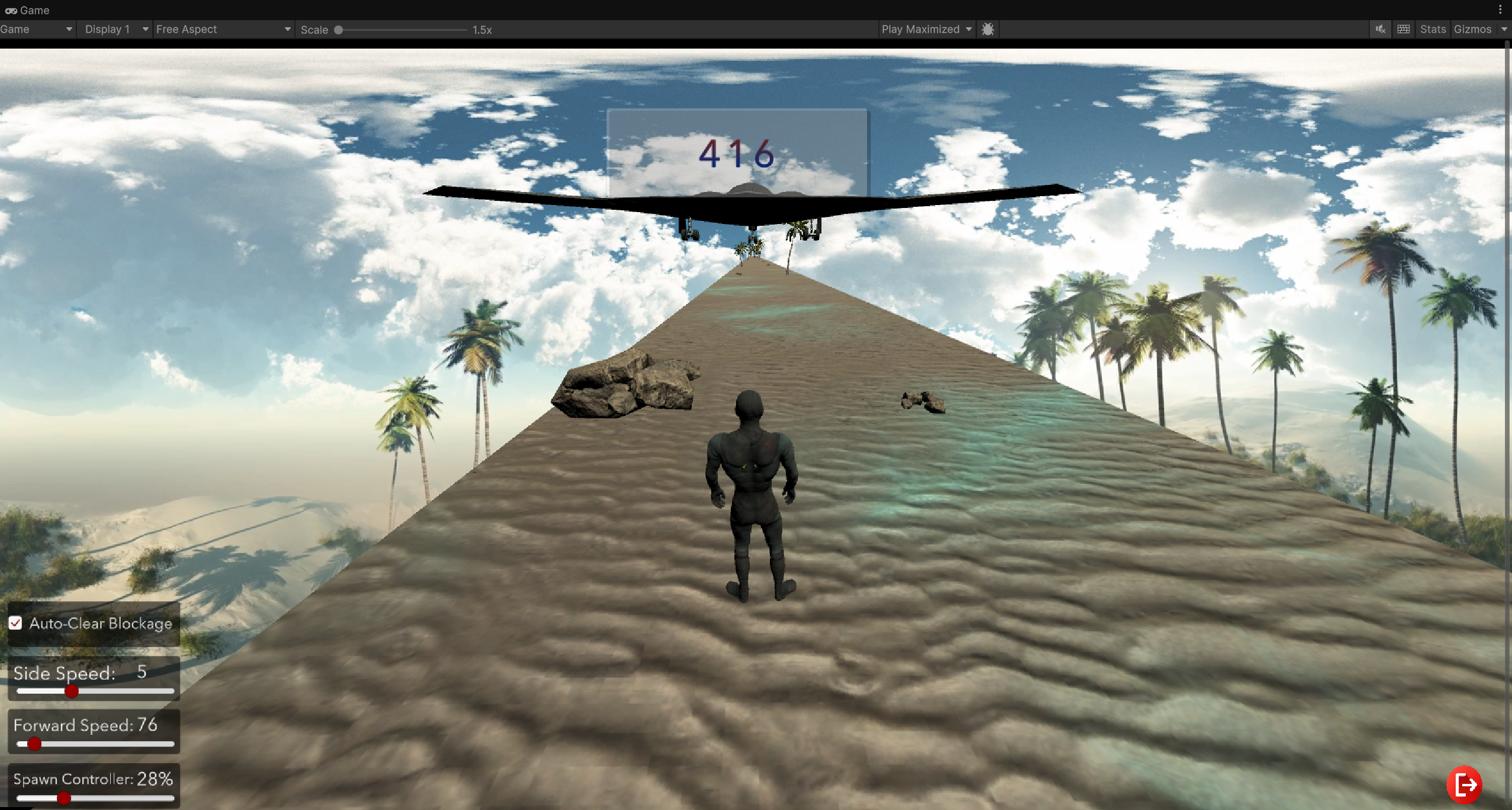}
    \caption{\textit{Momentum} during gameplay: the player advances along the procedurally streamed terrain while the aerial evaluation agent inspects the corridor ahead. The on-screen overlay exposes the runtime sliders for forward speed, side speed, and spawn density.}
    \label{fig:gameconcept}
\end{figure}

The generated environment is also the substrate for evaluation. The system not only produces new terrain and objects, but also examines whether the generated path remains traversable for the player. Autonomous agents are positioned ahead of the player to inspect the generated sections and report blocked or unsolvable scenarios.

\subsection{Player Physics}

The player physics script controls character motion along the forward (Z) and lateral (X) axes. The script follows a direct execution flow: inspector-exposed constants, component initialisation, frame updates, input handling, ground detection, and finally the velocity update applied to the Rigidbody (Unity's physics-driven motion component, which applies forces, gravity, and collisions to a game object). This structure is kept explicit because the player controller is the base movement layer on which the procedural generation and evaluation systems operate, and clarity at this layer simplifies reasoning about every system above it.

\subsubsection{Fields and Inspector Exposed Constants}

The \texttt{[SerializeField]} variables (script fields exposed to the Unity editor while remaining private to other code) are exposed to the Unity Inspector (the editor panel that displays and edits a game object's properties at design time) and attached to the attached to the player runner prefab. These values allow the player movement behaviour to be tuned directly from the editor without changing the script every time. The main exposed fields are:

\begin{itemize}[leftmargin=*]
    \item Kinematic constants: \texttt{runSpeed}, \texttt{sideSpeed}, \texttt{jumpForce}, and \texttt{jumpForwardBoost}.
    \item Visual movement parameters: \texttt{angleOfLean} and \texttt{speedOfLean}, used for smooth side movement during the run.
    \item Component references involving the \texttt{Rigidbody} and \texttt{Animator}.
\end{itemize}

\subsubsection{Initialisation}

The scene script starts with the \texttt{Start()} function calling the component initialisation logic. At this point, the Rigidbody configuration is applied so that the player does not rotate incorrectly during movement. Rotation is locked across the X and Z directions, linear damping is applied with \(c = 2f\), and interpolation is enabled to give smoother motion. The safe spawn position is also stored for later re-spawn, so that the player can be teleported back after falling below the valid playable area. The initial boolean flags are then set as \texttt{isGrounded = 1} and \texttt{isJumping = 0}.

\subsubsection{Frame Loop}

The frame loop is separated according to Unity's execution model. The \texttt{Update()} function (called once per rendered frame) is used for input, animation and ground checks, while \texttt{FixedUpdate()} (called at a fixed physics interval, independent of frame rate) is used for the physics update where the linear velocity is overwritten and the model tilt is corrected.

\begin{table}[H]
\centering
\caption{Player movement frame loop.}
\label{tab:player_movement_frame_loop}
\begin{tabular}{lll}
\toprule
Callback & Frequency of Calls & Functionality \\
\midrule
\texttt{Update()} & Every rendered frame & Captures input, animations and ground checks \\
\texttt{FixedUpdate()} & Fixed physics interval & Overwrites linear velocity and fixes model tilt \\
\bottomrule
\end{tabular}
\end{table}

\subsubsection{Interpreting Keyboard Input Details}

Holding \texttt{W} sets the player into the running state for the movement logic and animation controller. Holding \texttt{A} or \texttt{D} provides a discrete horizontal value for side movement. Tapping \texttt{Space} while the player is grounded sets the \texttt{isJumping} flag to true. This keeps the control scheme simple and matches the endless-runner structure of the game.

\subsubsection{Ground Detection Technique}

Ground detection is handled using ray casting (projecting an invisible line from a point in a chosen direction and reporting the first surface it intersects). A short ray of 0.3 metres is fired from the feet of the player. If this ray hits the \texttt{groundMask} layer, then \texttt{isGrounded} becomes true and the landing state is cleared. This check is required before applying the jump logic, because the player should not be allowed to repeatedly jump while already in the air.

\subsubsection{Dynamics Update}

Forward velocity is not applied as an instantaneous change. The script uses linear interpolation~\cite{unity2025mathflerp} so that the current forward velocity advances gradually towards the target, producing a smooth transition rather than a step:

\begin{equation}
v_z \leftarrow \operatorname{lerp}(v_z,\, v_{\mathrm{target}},\, 10\,\Delta t),
\end{equation}

where

\begin{equation} v_{\mathrm{target}} = \begin{cases} \texttt{runSpeed}, & \text{running and grounded (direct assignment),} \\ 0, & \text{grounded but not running (lerp target).} \end{cases} \end{equation}

When the player is running and grounded, \(v_z\) is assigned \texttt{runSpeed} directly. The interpolation in equation~(1) applies only when the player decelerates to zero.

For small \(\Delta t\), the recurrence approximates exponential approach,

\begin{equation}
v_z(t) \approx v_z(0)\,e^{-10t},
\end{equation}

so the controller accelerates and decelerates with a time constant of approximately 0.1\,s. Here, \texttt{runSpeed} is the configured running speed, \(v_z\) the current forward velocity, \(\Delta t\) the physics timestep, and the factor 10 controls the rate at which the velocity approaches its target.

Lateral motion is computed from the discrete horizontal input:

\begin{equation}
v_x = h \cdot \texttt{sideSpeed},
\end{equation}

where \(h \in \{-1, 0, +1\}\), with \(h=-1\) for left, \(h=+1\) for right, and \(h=0\) when no lateral key is pressed. The jump action overwrites the vertical component using the configured jump force,

\begin{equation}
v_y \leftarrow j,
\end{equation}

and, when the player is already running, an additional forward impulse is applied along the Z-axis,

\begin{equation}
v_z \leftarrow v_z + \texttt{jumpForwardBoost} \cdot \mathbb{1}_{\mathrm{running}},
\end{equation}

where \(\mathbb{1}_{\mathrm{running}}\) equals 1 when the player is in the running state and 0 otherwise. After lift-off, the vertical position follows the standard kinematic equation,

\begin{equation}
y(t) = y_0 + v_y t - \tfrac{1}{2}\,g\,t^2, \qquad g \approx 9.81\,\mathrm{m/s^2},
\end{equation}

with \(y_0\) the vertical position at the start of the jump, \(y(t)\) the position at time \(t\), and \(v_y\) the vertical velocity immediately after lift-off. A linear damping coefficient of \(c = 2\,\mathrm{s^{-1}}\) is applied through the Rigidbody, so that any unforced velocity decays as \(v(t) = v_0 e^{-2t}\). This prevents residual drift after input release without producing an abrupt stop. If the player falls below \(y = -8\,\mathrm{m}\), the controller triggers a respawn that teleports the character back to the configured spawn point and resets the velocity. Up to nine respawns are permitted per run before the game-over state is entered. This safeguard is required because the procedural ground occasionally contains transient gaps between tile boundaries during navigation rebakes.

\subsection{Procedural Terrain Generation}

The procedural terrain generation system in \textit{Momentum} provides the player with a continuous traversable surface, while managing object spawning, navigation-mesh updates, and tile lifecycle during runtime. The system is designed to be scalable and memory-efficient. The terrain is not produced through noise-based height maps, since the game uses uniform ground tiles where height variation is minimal. Emphasis is therefore placed on continuous tile streaming, controlled object placement, navigation surface management, and incremental cleanup, rather than on topographical synthesis.

\subsubsection{Core Script Parameters}

The terrain generation script is controlled through the following core parameters:

\begin{itemize}[leftmargin=*]
    \item Ground Tile Length: \(L_{\mathrm{tile}} = 96\,\mathrm{m}\), where the tiles are spaced along the Z-axis one after another.

    \item Navigational Mesh Coverage: Ahead: \(d_{ahead} = 600\,\mathrm{m}\), Behind: \(d_{behind} = 50\,\mathrm{m}\).

    \item Spawn Rate for Ground: \(\Delta t_{\min}\) and \(\Delta t_{\max}\), which define the shortest and longest time interval allowed between two consecutive tile spawns.

    \item Skybox Variants: Tile appearance is chosen based on the environment sky theme, using an index variable for tracking \texttt{skyboxVariantIndex}.
\end{itemize}

\subsubsection{Tile Spawning}

Tiles are spawned according to two conditions. The first is a temporal trigger: a new tile is generated when the elapsed time since the previous spawn exceeds the active interval,

\begin{equation}
t > t_{\mathrm{lastSpawn}} + \Delta t_{\mathrm{spawn}}.
\end{equation}

The interval itself contracts as the player's speed increases,

\begin{equation}
\Delta t_{\mathrm{spawn}} = \operatorname{clamp}\bigl(\Delta t_{\max} - \gamma\, v_p(t),\ \Delta t_{\min},\ \Delta t_{\max}\bigr),
\end{equation}

with the player's instantaneous speed estimated from successive forward positions,

\begin{equation}
v_p(t) = \frac{|z_p(t) - z_p(t-\delta t)|}{\delta t}.
\end{equation}

The scaling constant is \(\gamma = 0.01\), and the clamp~\cite{unity2025mathfclamp} bounds the interval within \([\Delta t_{\min}, \Delta t_{\max}]\) for any input.

The second condition is a proximity check. When the most recently generated tile is sufficiently far ahead of the player, spawning is suspended,

\begin{equation}
z_g > z_p + 1000 \;\Rightarrow\; \text{spawning suspended,}
\end{equation}

so that the system does not extend the world beyond the range that any agent or the player will reach in the near term. The parameters above denote: \(t\) the current time; \(t_{\mathrm{lastSpawn}}\) the time of the previous spawn; \(\Delta t_{\mathrm{spawn}}\) the active spawn interval; \(v_p(t)\) the player's instantaneous speed; \(z_p(t)\) the player's forward position; \(z_g\) the forward position of the most recent tile; \(\gamma\) the speed-to-interval coupling constant; and \(\delta t\) the sampling step used to estimate \(v_p\).

\subsubsection{Positioning of the Tiles}

When a new ground tile is spawned, its Z-axis position is calculated as:

\begin{equation}
z_{g,\mathrm{new}} = z_g + L_{\mathrm{tile}}.
\end{equation}

where \(z_g\) is the Z-axis position of the last tile which has been spawned. All the tiles spawned are instantiated as child objects of the game object called \texttt{GroundContainer} to organise the hierarchy. After the instantiation, the environment objects are spawned via the object spawn manager.

\subsubsection{Navigational Mesh Management}

The navigation surface, also called the NavMesh (a simplified walkable map that Unity automatically generates from scene geometry so AI characters can pathfind across it), is the substrate on which the AI agents move~\cite{unity2025navmesh}. It can be understood as a runtime traversability map that defines where an agent may walk, jump, or stop. Because the ground is generated dynamically, the navigation surface must also be updated during runtime; a single bake at scene load is not viable.

Three triggers initiate a re-bake:

\begin{itemize}[leftmargin=*]
    \item \textbf{Time-based:} every \(T_{\mathrm{navMesh}} = 1\,\mathrm{s}\).
    \item \textbf{Count-based:} every \(N = 5\) newly spawned tiles, controlled through the configured bake interval, or whenever the player advances by more than two tile lengths \((2 L_{\mathrm{tile}})\).
    \item \textbf{Position-based:} whenever the player moves more than 50\,m since the last bake.
\end{itemize}

The baked region is re-centered on the player at every rebuild,

\begin{equation}
z_{\mathrm{centre}} = z_p + \frac{d_{\mathrm{ahead}} - d_{\mathrm{behind}}}{2},
\end{equation}

so that its coverage is the interval \([z_p - d_{\mathrm{behind}},\ z_p + d_{\mathrm{ahead}}]\), with \(d_{\mathrm{ahead}}=600\,\mathrm{m}\) and \(d_{\mathrm{behind}}=50\,\mathrm{m}\).

The navigation rebake (re-computing the walkable map from the current geometry) is performed asynchronously (off the main thread, so the game does not freeze while it runs) to avoid frame drops (single rendered frames that take noticeably longer than the budget and produce a visible stutter). An earlier synchronous implementation produced visible stalls~\cite{unity2025profiler} because the engine halted the main thread at every rebuild while it walked the colliders, surfaced the mesh, and uploaded it. The current implementation issues the rebuild as a non-blocking job, summarised in Algorithm~\ref{alg:async_navmesh_rebuild}, so that the navigation surface remains current without forcing the main loop to halt during a bake.

\begin{algorithm}[H]
\caption{Asynchronous navigation-mesh rebuild.}
\label{alg:async_navmesh_rebuild}
\begin{algorithmic}[1]
\State Wait until any in-flight bake job has completed.
\State Collect navigation-mesh sources from the active scene.
\State Recompute the bake bounds around the current player position.
\State Issue a non-blocking navigation update using the existing bake data, build settings, collected sources, and bounds.
\State Yield until the bake job reports completion.
\end{algorithmic}
\end{algorithm}

\subsubsection{Clean-up of Procedural Terrain}

For smooth performance and optimal use of memory, every generated ground tile checks the player's current position and destroys itself after the player has safely moved ahead. This is required because the terrain is generated continuously, and keeping every old tile in the scene would slowly increase the number of active objects without adding anything useful to the current gameplay.

Let \(z_{\mathrm{tile}}\) be the Z-axis position of the tile and \(z_p\) be the coordinate of the player on the Z-axis. For every single tile, a threshold is set as:

\begin{equation}
z_{\mathrm{threshold}} = z_{\mathrm{tile}} + 100.
\end{equation}

The moment the player moves 400 units beyond this value, the tile is scheduled to delete itself:

\begin{equation}
z_p > (z_{\mathrm{tile}} + 100) + 400 = z_{\mathrm{tile}} + 500.
\end{equation}

The implementation uses a delayed destruction call:

\begin{verbatim}
Destroy(this.gameObject, 5.0f)
\end{verbatim}

The 5-second delay works as a small safety buffer for remaining references before Unity removes the object from the scene. This keeps the older procedural terrain from staying in memory once it is no longer part of active gameplay, without making the ground disappear too aggressively.

\subsection{Wave Function Collapse-Inspired Object Spawning}

Wave Function Collapse is used in this project as an object-spawning method rather than a complete map-generation technique. The implementation does not use neural networks or a trained machine learning model. Instead, a simple one-dimensional WFC-inspired process is used to spawn environmental objects on the generated ground tiles. The aim is to keep the spawning process random enough for variation, while still applying basic constraints so that objects do not appear in completely uncontrolled positions.

The \texttt{EnvironmentObjectSpawnManager} script controls the spawning, adaptation and lifecycle of obstacle prefabs (pre-configured Unity templates that can be instantiated repeatedly as identical copies at runtime) that are already stored as assets. The object prefabs are organised in variants based on the current skybox theme and are placed in the folder structure:

\begin{verbatim}
PF <variant> <object>
\end{verbatim}

During initialisation, all prefabs are cached by the theme of the environment using \texttt{BuildPrefabCache()}. This means that when the skybox changes, the spawning system can select objects from the corresponding variant while still keeping fallback conditions in place.

\subsubsection{Grid Construction}

Each generated ground tile is divided across the X-axis into a one-dimensional grid. This grid is used to decide the possible positions where environmental objects can be placed. The width of the available spawning region is first calculated as:

\begin{equation}
W = x_{\max} - x_{\min}.
\end{equation}

The number of cells for the tile is then taken as the maximum between the rounded-up width and a minimum value of 12:

\begin{equation}
N = \max\bigl(\lceil W \rceil,\, 12\bigr).
\end{equation}

This same calculation is implemented in the script as:

\begin{verbatim}
float width = xAxisRange.y - xAxisRange.x;
int cells = Mathf.Max(Mathf.CeilToInt(width), 12);
\end{verbatim}

For every tile, the number of spawning attempts is calculated from the spawn-density percentage. This makes the amount of generated objects dependent on the configured spawn rate, while still keeping it bounded by the number of available grid cells:

\begin{equation}
K = \operatorname{clamp}\!\left(\left\lfloor N \cdot \frac{p_{\mathrm{spawn}}}{100} \right\rfloor,\; 0,\; N\right).
\end{equation}

Here, \(p_{\mathrm{spawn}}\) represents the density percentage used for object spawning. The ground agent lane is also preserved using \texttt{clearHalfWidth}, so that the grid does not place objects directly inside the main path required by the player. This makes the spawning process controlled enough to avoid obvious blockage, but still random enough to keep the generated environment unpredictable.

\subsubsection{Placement}

The placement stage starts after the grid cells and the number of spawning attempts have already been calculated. The cells are randomly chosen, but the selection is not completely left open. Once a cell is selected, its neighbouring cells \((i-1, i+1)\) are marked as occupied. This prevents objects from clustering too closely on the same ground tile and keeps the placement more distributed across the available width.

The X-coordinate of the object is fixed from the selected grid cell. The Z-axis position is given a small jitter so that objects are not spawned in a perfectly straight line:

\begin{equation}
z = z_{\mathrm{ground,centre}} + \mathcal{U}(-z_{\mathrm{jitter}},\, z_{\mathrm{jitter}}).
\end{equation}

After this position is calculated, a ray cast is performed towards the ground. The purpose of this ray cast is to confirm that valid ground exists at the selected position. If the ray cast does not find a valid ground hit, that object is skipped. This prevents objects from being instantiated in empty space or outside the playable ground area.

The correct prefab is then selected from the current skybox variant. Every skybox variant has multiple prefabs associated with it, so the visual objects spawned in the environment remain connected with the active skybox theme. The prefab is selected randomly from the available objects of that variant.

Once the ray cast returns a valid ground hit, the object is instantiated at that point. Snapping to ground is performed through \texttt{SnapToGround()}, where the boundary of the object is used to align the base with the terrain. Some specific objects with complex geometry at their base are forced down into the Y-axis. These objects mainly involve variants of trees and a rock cliff. This workaround is necessary because these assets do not always have a stable lower base after spawning. Certain objects are also allowed to float above the ground where the visual effect is intentional.

\subsubsection{Despawning Objects}

The generated objects are removed using a separate despawning rule from the ground tiles. This is required because the environmental objects do not need to remain in the scene once the player has already passed them. If they are kept for too long, they continue to occupy memory and physics resources without contributing to the current game state.

For each spawned object, the system compares the Z-axis position of the player with the Z-axis position of the object. The object is removed when the player moves more than 50 units ahead of it:

\begin{equation}
z_{\mathrm{player}} - z_{\mathrm{object}} > 50.
\end{equation}

This threshold is deliberately smaller than the terrain clean-up threshold. Objects are therefore deleted before the ground tile itself is removed. This ordering is important because if the ground disappears first, the remaining objects may fall through the environment, accelerate due to gravity and remain active as unnecessary physics bodies.

The despawning mechanism therefore keeps the generated scene bounded around the active play area. It allows the game to continue producing new objects ahead of the player while removing old objects behind the player before they become an avoidable runtime cost.

\subsection{Skybox and Environment Variation}

The skybox and environment-variation system provides visual change as the player progresses along the positive Z-axis. This is important in an endless-runner setting because the player is not moving towards a fixed endpoint. The surrounding environment therefore needs to change over time so that the generated world does not appear visually static.

In \textit{Momentum}, this behaviour is managed through the \texttt{SkyboxChanger} system. The system updates the active skybox material and the associated 360-degree environment video during runtime. The transition is driven by player progression: the system monitors the player's Z-axis position and initiates a change when the player crosses the next threshold, incremented by \(\Delta z = d_{\mathrm{interval}}\), where \(d_{\mathrm{interval}}=2000\) is the configured distance interval

The available skybox videos are stored as indexed assets. A stack mechanism tracks recently displayed variants so that the same visual state is not selected repeatedly. Once all variants have been displayed, the stack is reset and the selection cycle continues. This gives the environment a controlled form of variation without requiring a fully handcrafted sequence of visual states.

The transition itself is handled through exposure control. The current skybox fades out until the exposure reaches zero. At that point, the new material and video clip are assigned while the scene is still dark. The new skybox then fades back in to the intended exposure value. This avoids abrupt frame changes and visible flashes during the swap. The function \texttt{DynamicGI.UpdateEnvironment()} is also called during the transition so that the scene lighting remains consistent with the active skybox.

The same variation is reflected in procedural object spawning. Environmental prefabs are grouped according to skybox variants, so the active skybox also influences which objects can appear on the generated ground. This links the background, lighting and spawned objects into the same environment theme while leaving the terrain-generation logic unchanged.

\subsection{Runtime User Interface}

The runtime user interface acts as a small experimental control layer inside the game. It is not only a display element, but also a way to tune important game parameters during execution. This is useful because the project depends on procedural generation and runtime evaluation, where speed and object density can change how difficult or stable the generated path becomes.

The interface is implemented using Unity's Canvas UI system~\cite{unity2025docs} and TextMeshPro elements. Three main parameters are exposed through sliders: player forward speed, player side speed and environment object spawn rate. The forward-speed slider changes the running speed of the player. The side-speed slider changes the lateral movement speed, but within a smaller boundary so that the player does not easily leave the valid ground area. The spawn-rate slider modifies the object spawn percentage and updates the corresponding on-screen label during gameplay.

The score is treated as a progression measure rather than a separate reward system. Since the player advances along the Z-axis, the score is calculated from the player's Z-axis position with an offset added for the initial spawn position:

\begin{equation}
S(t) = \max\!\bigl(0,\; z_p(t) + \Delta_{\mathrm{offset}}\bigr).
\end{equation}

Here, \(z_p(t)\) is the player position on the Z-axis at time \(t\), and \(\Delta_{\mathrm{offset}}=45\). The offset is used because the player starts from \(-45\,\mathrm{m}\) on the Z-axis during initialisation. The score is updated every frame and displayed through the TextMeshPro UI element.

\section{Analysis and Evaluation}

\subsection{Aerial Agent}

The aerial agent is used as the first runtime evaluator of the generated path. Since the terrain and obstacles are produced during play, the game cannot assume that the next section of the level will always remain traversable. The agent therefore moves ahead of the player and inspects the upcoming region before the player reaches it. In this sense, the aerial agent works as an early warning layer for the procedural generation system.

Unlike the player, the aerial agent is not meant to interact with the terrain as a physical character. It observes the generated corridor from above and checks for problematic object placement, blocked sections, or situations where the generated space may become invalid. This is useful because the agent can evaluate the level without being slowed down by normal ground movement, collisions, or gravity.

\subsubsection{Agent Trajectory}

At each frame, the aerial agent calculates a target position ahead of the player. Since \(\Delta_z\) represents a fixed forward offset, the target Z-position is calculated by adding this offset to the player's current Z-axis position and then clamping it to the farthest generated ground position:

\begin{equation}
z_{\mathrm{aerial,target}} = \min(z_p + \Delta_z,\; z_{\mathrm{ground,last}}).
\end{equation}

Here, \(z_p\) is the player's position on the Z-axis, \(\Delta_z\) is the fixed forward offset used by the aerial agent, and \(z_{\mathrm{ground,last}}\) is the position of the farthest recently spawned ground tile. The minimum operation prevents the aerial agent from moving into empty space beyond the generated terrain.

The full target position is represented as:

\begin{equation}
p_{\mathrm{aerial,target}} =
\begin{bmatrix}
x_p + \Delta x \\
h_f \\
z_{\mathrm{aerial,target}}
\end{bmatrix}.
\end{equation}

where \(x_p\) is the player's X-axis position, \(\Delta x\) is the lateral offset, and \(h_f\) is the fixed altitude maintained by the aerial agent. The aerial agent is kept independent from the engine's gravity and normal physics response because its role is to inspect the upcoming generated path, not to behave as another physically simulated character.

\subsubsection{Speed Control}

The position of the aerial agent is updated using a smoothed-velocity interpolation routine. Instead of placing the aerial agent directly at the target position, the method moves it towards the target through a smoothed velocity update:

\begin{equation}
p_f(t+\Delta t) = \operatorname{SmoothDamp}(p_f(t),\, p_{\mathrm{aerial,target}},\, v_f,\, T_s,\, v_{\max}).
\end{equation}

Here, \(p_f(t)\) is the current aerial agent position, \(p_{\mathrm{aerial,target}}\) is the target position calculated ahead of the player, \(v_f\) is the current velocity reference used by the smoothing function, \(T_s\) is the smoothing time, and \(v_{\max}\) is the maximum speed allowed for the aerial agent.

This keeps the agent responsive, but prevents sudden jumps in its movement. The aerial agent follows the generated path ahead of the player while remaining constrained by the smoothing and maximum-speed parameters. This is important because the agent is used for evaluation, so unstable or abrupt motion would make the scanning process less reliable.

\subsection{Ray Casting}

Ray casting is used by the aerial agent to evaluate whether the procedurally generated path remains safe and usable. Since the environment is created during runtime, the generated space must be checked before the player reaches it. The aerial agent performs this check by scanning the terrain ahead using downward rays and volumetric physics sweeps.

The purpose of ray casting in this project is not only to detect whether ground exists. It is also used to identify blocked regions, unsafe object placement and cases where the player may not have enough continuous space to move forward. A single ray hit is therefore not enough to decide that a region is playable. The scan needs to consider the width of the path and the presence of blocking colliders across that width.

\subsubsection{Segmentation and Tile Probing}

The area ahead of the aerial agent is divided into tile-like segments along the Z-axis. Each segment is scanned using horizontal probe rows, with multiple ray origins distributed across the width of the ground. From each origin, a ray is fired downward towards the generated terrain:

\begin{equation}
\mathrm{origin} = (x,\, y_{\mathrm{scan}},\, z_{\mathrm{row}}).
\end{equation}

Here, \(x\) varies across the width of the ground, \(y_{\mathrm{scan}}\) is the height from which the ray is cast, and \(z_{\mathrm{row}}\) is the Z-axis position of the row currently being examined.

A row is considered passable only when it contains a continuous clear interval of sufficient width for the player to move through:

\begin{equation}
\exists\, [a,b] \;\text{such that}\; (b-a) \geq w_{\mathrm{clear}} \;\land\; C(x)=0 \;\;\forall x \in [a,b].
\end{equation}

In this condition, \([a,b]\) represents a continuous interval across the scanned row, \(w_{\mathrm{clear}}\) is the minimum required clear width, and \(C(x)=0\) indicates that no blocking collider is detected at the sampled position. If no such interval exists, the row is treated as unsafe.

This formulation evaluates the path as a traversable corridor rather than as isolated ground hits. A generated section is therefore accepted only when there is enough uninterrupted space for the player to pass through. The volumetric checks described in the next subsection extend this test by detecting physical collider volume that may not be fully captured by individual rays.

\subsubsection{Volumetric Sweeps}

Ray casting provides a narrow, line-based view of the generated corridor. This is useful for sampling the ground and checking whether a clear passage exists, but it does not fully describe the physical volume occupied by irregular objects. For this reason, the scanner supplements the ray probes with \texttt{Physics.OverlapBox} checks (queries that ask the physics engine which colliders intersect a defined 3D box, rather than only what a single line hits).

The overlap-box test is centred on the tile being inspected and covers the traversable corridor window. Any collider intersecting this volume is captured as part of the scan result. This makes the evaluation less dependent on whether a thin ray happens to strike the exact obstructing part of an object. It is particularly relevant for prefabs with uneven geometry, extended branches, wide bases, or collider shapes that do not align neatly with the visual mesh.

In this form, the ray probes and volumetric sweeps serve different purposes. The rays estimate whether a continuous path exists across the scanned row, while the overlap-box check verifies whether the same corridor is physically occupied by colliders. The two checks together give a more reliable assessment of whether the generated region is actually usable by the player.

\subsubsection{Identification and Automatic Removal}

Objects detected through the ray probes or overlap-box checks are recorded using their collider information and the name of the corresponding root object. This keeps the blockage report tied to the actual generated prefab rather than only to an anonymous physics hit. The reported data therefore remains useful for later analysis, because the problematic object can be traced back to the spawned environment asset.

When the automatic-removal option is enabled, the scanner can destroy objects that block the corridor before the player reaches that section. This behaviour is kept as part of the evaluation mechanism because the generated world is produced at runtime, and some invalid object placements may need to be handled immediately rather than only reported after failure.

Even when an object is removed, the blockage is still recorded with its relevant details, including object identity, size and layer information. The removal step therefore does not hide the generation problem. It only prevents the detected blockage from becoming an immediate obstacle for the player, while preserving the evidence needed to study why that object placement was unsafe.

\subsection{Ground Agent}

The ground agent provides a second form of validation from the ground level. While the aerial agent examines the generated path from above, the ground agent tests whether the same terrain can be traversed through the runtime navigation surface. This distinction is important because a region may appear open from an overhead scan, but still be difficult for a ground-based agent to navigate because of NavMesh boundaries, collider placement or discontinuities between generated tiles.

The ground agent operates ahead of the player and follows the generated path as a traversable agent. Its role is to identify real-time navigation issues that are closer to the player's actual movement conditions. In this sense, it acts as a ground-level counterpart to the aerial agent. The aerial agent checks the corridor geometrically, while the ground agent checks whether the procedural terrain remains navigable through the same navigation structure used for agent movement.

This gives the evaluation system two complementary views of the generated content: one based on spatial scanning and another based on practical path traversal. Together, they reduce the chance that a generated section is accepted only because it satisfies one form of inspection.

\subsubsection{Navigation and Behaviour}

The ground agent validates the generated terrain from a ground-navigation perspective. The aerial agent can inspect the corridor from above, but an overhead scan does not confirm that the same region is navigable through the baked NavMesh. The ground agent therefore acts as a second evaluator by attempting to move through the generated path using the runtime navigation system.

The target position of the ground agent is calculated ahead of the player:

\begin{equation}
p_{\mathrm{ground,target}} = \begin{bmatrix} x_p \\ y_{\mathrm{navMesh}} \\ z_p + D_{\mathrm{lookAhead}} \end{bmatrix}.
\end{equation}

Here, \(x_p\) is the player's X-axis position, \(y_{\mathrm{navMesh}}\) is the valid height sampled from the navigation mesh, and \(D_{\mathrm{lookAhead}}\) is the forward distance used to keep the ground agent ahead of the player. The target is therefore tied to the navigable surface rather than being an arbitrary point in world space.

The ground agent speed is adjusted according to the player's forward velocity:

\begin{equation}
v_{\mathrm{runner}} = \max(10, |v_{p,z}|) + v_{boost}.
\end{equation}

The lower bound of 10 ensures that the ground agent continues moving even when the player's forward velocity is low. The additional boost keeps the ground agent separated from the player, allowing navigation problems to be detected before they become immediate gameplay failures.

\subsubsection{Movement Stuck and Recovery}

A practical issue observed during runtime was that the ground agent could become stuck around overlapping NavMesh regions between procedurally generated ground tiles. This failure is different from an ordinary obstacle blockage. The path may appear open in the scene, but small discontinuities or boundary artefacts in the baked navigation surface can prevent the agent from progressing.

The stuck condition is evaluated using both movement progress and NavMesh velocity. The ground agent is treated as stuck when its Z-axis position changes by less than the configured threshold while its NavMesh velocity is nearly zero. Using both conditions reduces false positives, since the agent may briefly slow down or adjust its direction without actually being trapped.

When this condition is detected, the recovery mechanism moves the ground agent forward by the configured step, places it on the nearest valid NavMesh position, resets the current path, and applies a cooldown before another recovery attempt can occur. The cooldown prevents repeated recovery actions during a short navigation disturbance.

This recovery mechanism does not remove the underlying generation issue. Its purpose is to keep the evaluator active when the failure is caused by minor NavMesh overlap or tile-boundary artefacts, so that the system can continue inspecting the generated path ahead.

\subsubsection{Corridor Scanning}

The ground agent also performs corridor scanning and obstacle detection, but from a ground-navigation perspective. While the aerial agent scans the generated region from above, the ground agent checks whether the same region can be approached through the baked NavMesh surface.

This gives the evaluation system a second form of evidence. A section may pass the overhead scan because there is visible ground and some clear space, but the ground agent may still fail if the navigable surface is broken, blocked, or not connected properly across generated tiles. The ground agent scan therefore focuses less on visual openness and more on practical traversability.

Detected problems are treated as part of the runtime evaluation report. In this way, the ground agent acts as a ground-level validator for the procedural generation system, complementing the aerial agent rather than replacing it.

\subsubsection{Obstacle Layer, Volumetric Sweep, and Blockage Reporting}

The ground agent applies obstacle detection through a dedicated obstacle layer rather than treating every collider in the scene as a possible blockage. The agent's \texttt{obstacleMask} (a bit mask telling the physics queries which Unity layers count as obstacles and which to ignore) is configured to the \texttt{Obstacles} layer in the Unity Inspector, and the obstacle prefabs used by the procedural generator are marked with this layer. This keeps the scan focused on objects that can actually interfere with traversal, instead of mixing terrain, decorative objects, agent colliders and unrelated scene elements into the same detection process.

The ground agent uses the same general scanning idea as the aerial agent, but its interpretation is different. The aerial agent can remove blocking objects when automatic removal is enabled, whereas the ground agent does not modify the generated scene. Its role is diagnostic. It performs ray-based checks and supplements them with \texttt{Physics.OverlapBox} sweeps for irregularly shaped objects, then records the detected obstacle colliders and the corresponding tile position.

This distinction is useful for evaluation. If the ground agent reports a blockage, the result represents a ground-navigation failure rather than an automatically corrected generation issue. The blockage data is then passed to the reporting pipeline, where it can be preserved for later inspection instead of being lost during runtime.

\subsection{Crash Report}

The crash-report system is the persistent record of the evaluation process. Since procedural generation can fail because of a particular combination of player state, object density, tile position, skybox variant or scanner configuration, the report cannot only state that the path was blocked. It needs to preserve the surrounding conditions that produced the blockage.

The system therefore captures, combines and exports detected corridor blockages during the game session. In this project, the report works both as immediate feedback inside the game and as a post-gameplay artefact for analysing why a generated section became unsafe.

\subsubsection{Blockage Detection and Data Capture}

When an impassable tile is detected, the blockage information is sent through the \texttt{BlockageReporter} class. Each report is stored as a serialisable data object (a plain data record that Unity can convert into bytes so it survives a scene change or can be written to disk), so that the failure state can be carried across scenes and later rendered in the exit-report view. The collected reports are stored in a static cache, which allows the detected blockage information to survive the transition from the running game scene to the report scene.

The captured fields are summarised in Table~\ref{tab:blockage_report}.

\begin{table}[t]
\centering
\caption{Crash-report data captured for procedural blockage analysis.}
\label{tab:blockage_report}
\begin{tabular}{ll}
\toprule
Category & Captured Data \\
\midrule
Game context & Scene name, timestamp \\
Player state & Player position, speed \\
Environment state & Skybox variant, latest ground position, tile length \\
Generation parameters & Spawn percentage, X range, clear width, jitter, Y offset \\
Scanner context & Tile position, probe position, ray spacing, hit count \\
Blocking objects & Object name, position, size, layer \\
\bottomrule
\end{tabular}
\end{table}

This structure keeps the report close to the actual generation process. The object name identifies the spawned prefab involved in the blockage, while the scanner and generation parameters show the procedural conditions under which the blockage was detected.

\subsubsection{Exit Scene Visualisation}

The exit-report scene renders the stored blockage data using TextMeshPro. The \texttt{BlockageReportDisplay} component converts the cached blockage records into a readable report, with formatting and margins suitable for viewing inside the game. The scene therefore acts as a bridge between runtime detection and human inspection.

If no blockage data is available, the scene does not remain blank. A fallback message is displayed instead, indicating that no reports are present and giving possible reasons such as no detected blockages, data loss or cache reset. This makes the report scene useful both when failures are detected and when the evaluation run completes without recorded blockage data.

\subsubsection{PDF Export Functionality}

The PDF export mechanism provides a persistent version of the crash report outside the Unity scene. The \texttt{PdfExporter} script collects the report text and writes it into a readable PDF document. The exporter handles line wrapping, pagination, margins and report formatting, which is necessary because blockage reports can vary in length depending on the number of detected objects and scanner entries.

The implementation writes the PDF structure directly, including pages, font objects and text streams~\cite{adobe2008pdf}. The resulting document can be saved locally in a stand-alone build. This makes the crash report usable beyond the immediate game session and allows the detected procedural failures to be reviewed after execution.

\subsection{Quantitative Evaluation}
\label{sec:quant-eval}

The evaluation framework defined in Section~\ref{sec:eval-questions} admits two classes of result. \emph{Structural} results follow analytically from the parameters fixed in the spawner and scanner scripts and are reported below as derived numbers. The remaining metrics ($R_{\mathrm{block}}$, $R_{\mathrm{remove}}$, $\rho_{\mathrm{spawn}}$, frame-time distribution, prefab entropy, and Jaccard overlap) require runtime instrumentation; their empirical instantiation on the released build is left as future work and is described as a limitation in Section~\ref{sec:limitations}.

\subsubsection{Structural results derived from code}
\label{sec:struct-results}

Three structural properties of the system are determined entirely by the constants in the spawner and scanner scripts and require no runtime measurement.

\paragraph{F1 --- Spawner saturation bound (supports H2).}
With the X-range $[-7.1, 10.55]$ giving $W = 17.65~\mathrm{m}$ and $N = \max(\lceil W\rceil, 12) = 18$ cells, a clear half-width of $2~\mathrm{m}$ excludes the four cells whose centres satisfy $|x_i| < 2$, leaving $14$ candidate cells. Because each successful placement marks both itself and its two neighbours as occupied (Section~4.5.2), the maximum number of objects that can ever coexist on a single tile is the size of a maximum independent set on the path of $14$ available cells, which is $8$. Solving $\lfloor N\cdot p_{\mathrm{spawn}}/100\rfloor = 8$ gives a saturation threshold $p^{\star}_{\mathrm{spawn}} \approx 44\%$. The spawn-density slider therefore exposes a controllability ceiling that is structural: requesting $50$\%, $75$\%, or $100$\% cannot place more objects than requesting $\approx 44$\%, and any extra demand is dissipated as failed re-roll attempts under the safety counter \texttt{attempts < cells * 5}.

\paragraph{F2 --- Per-segment scan cost (supports H3).}
A single call to \texttt{FlyerCorridorScanner.ScanCorridor} performs $\lceil L_{\mathrm{seg}}/g_z\rceil + 1 = 21$ horizontal probe rows over a segment of length $L_{\mathrm{seg}} = 10~\mathrm{m}$ with row gap $g_z = 0.5~\mathrm{m}$. Each row sweeps the corridor with X-step $0.05~\mathrm{m}$ across a width of $2.15~\mathrm{m}$, giving up to $\lfloor 2.15/0.05\rfloor + 1 = 44$ rays in the worst case (no early-exit on a sufficiently large clear gap). The per-segment ray budget is therefore bounded above by $21 \times 44 = 924$ rays, plus a single \texttt{Physics.OverlapBox} call. Because the scanner is gated by \texttt{tileZ <= \_lastScannedTileZAxis}, this work is performed at most once per $L_{\mathrm{seg}}$ of forward travel, irrespective of frame rate. The ground tile length is $96~\mathrm{m}$, so each ground tile incurs $\lceil 96/10\rceil = 10$ scanner segments. Note that the scanner segment length ($10~\mathrm{m}$) is not the same as the ground tile length ($96~\mathrm{m}$); the two operate at different granularities, and conflating them overestimates per-tile work by an order of magnitude.

\paragraph{F3 --- NavMesh coverage relative to look-ahead (supports H4).}
The aerial agent maintains a forward offset of $\Delta z = 200~\mathrm{m}$, while the ground agent uses a NavMesh look-ahead of $D_{\mathrm{lookAhead}} = 300~\mathrm{m}$, both within the baked region of $[z_p - 50, z_p + 600]~\mathrm{m}$ (Section~4.4.4). The ground agent therefore evaluates a larger forward window than the aerial agent under default settings, so the two agents do not in general cover the same set of tiles even when both are active.

Together, F1 establishes that the spawn-density slider has a controllability ceiling at $p^{\star}_{\mathrm{spawn}}\approx 44\%$, F2 establishes that scanner cost is constant per metre travelled rather than per frame and is bounded above by $924$ ray probes plus one \texttt{Physics.OverlapBox} call, and F3 establishes that the two agents inspect non-coincident forward windows under default settings. These are the claims that follow directly from the implementation; the corresponding empirical quantities ($\rho_{\mathrm{spawn}}$, mean frame time and its violation rates, $J(A,G)$) require a runtime-measurement batch and are not reported here.

\subsubsection{Threats to validity and limitations}
\label{sec:limitations}

The structural results F1--F3 follow from the code constants and therefore generalise only to the parameter set used in the released build; changing $W$, $N$, the clear half-width, the row gap, or the X-step changes both the saturation point and the ray budget. The remaining metrics defined in Section~\ref{sec:eval-questions} require runtime instrumentation and are not reported in this paper; their empirical instantiation on the released build is left as future work, together with the player study that would be required to extend the analysis from technical playability to human enjoyment. Ground-truth labels for false positives and false negatives are not directly observable at runtime and would rely on manual review of logged replays, which would introduce an annotator dependency. Commercial titles such as \emph{No Man's Sky}~\cite{mcaloon2016nomanssky} and \emph{Caves of Qud}~\cite{bucklew2017qud,bucklew2019qud} are retained only as qualitative inspiration for automated probing of generated content, not as numerical baselines, because their scale, asset budget, and production goals differ substantially from a single-author research prototype and a side-by-side comparison would not be meaningful.

\subsubsection{Connection to the canonical PCG evaluation taxonomy}
\label{sec:taxonomy}

The framework in Section~\ref{sec:eval-questions} instantiates three of the four axes identified by Cook et al.~\cite{cook2024evaluation} for procedural-level-generation systems: \emph{playability} via $R_{\mathrm{block}}$ and $R_{\mathrm{remove}}$; \emph{controllability} via the spawn-density response in conjunction with the structural bound F1; and \emph{performance} via the frame-time distribution against Unity's documented budgets~\cite{unityProfileBestPractices,unityProfileBlog}. \emph{Diversity} is approached by the prefab entropy $H^{\mathrm{norm}}_{\mathrm{prefab}}$. Of these, the present work delivers the structural component of controllability (F1) and the structural component of performance (F2) as derivable numbers; the remaining instantiations are framework-level and await runtime measurement.

\section{Legal, Social, and Ethical Considerations}

A runtime generator inherits the licensing terms of every asset on which it operates. \textit{Momentum} draws exclusively from a curated, in-project asset set, and the placement rules involve no example-driven model, no training corpus, and no external generative component. The legal exposure is therefore limited to the standard concerns of asset provenance and licence compatibility, which apply equally to any procedurally driven game built on third-party content. Any future variant that replaces the rule-based spawner with a learned model would be required to re-examine these conditions, particularly with respect to training data and reproductive output.

Two player-experience concerns warrant attention. The first is that endless-runner games depend structurally on repetition and reward progression, which can encourage prolonged play sessions~\cite{koepp1998dopamine}; this places a responsibility on the designer to ensure that engagement does not rely solely on compulsive repetition. The second is accessibility: a procedurally generated game should not assume uniform reaction time, visual capacity, or input preference across players. The runtime sliders for forward speed, lateral speed, and spawn density partially address this by giving the player a coarse means of adjusting difficulty without code-level changes.

Automated evaluation introduces additional responsibility. False negatives, in which an unsafe section is accepted as playable, terminate the run; false positives, in which a safe section is rejected, waste generated content and may alter what the player sees when automatic removal is enabled. Transparency therefore matters. The crash-report pipeline records every detected blockage with sufficient context to be inspected after the run, and human review of those records remains the appropriate basis for judging whether the generator's behaviour is acceptable. The agents reduce risk; they do not replace designer oversight.

\section{Conclusion}

This work presented \textit{Momentum}, an endless-runner game developed to study procedural content generation in conjunction with runtime evaluation. The project addressed two coupled concerns: the production of an unpredictable game environment during gameplay, and the verification that the generated content remains technically playable before it is encountered by the player.

The implementation combined continuous terrain streaming, constraint-driven object placement inspired by Wave Function Collapse, asynchronous navigation-surface management, runtime gameplay controls, autonomous aerial and ground evaluation agents, ray casting, volumetric physics sweeps, and structured crash reporting. Together these components form a pipeline in which content is not only generated, but also inspected before it reaches the player.

Two design decisions in the implementation merit emphasis. First, ray casting and volumetric physics sweeps perform complementary rather than overlapping functions: thin rays sample the corridor along its length, while volumetric checks capture the full collider footprint of objects whose geometry would otherwise slip between probe rows. Second, the aerial and ground agents inspect distinct properties of the same region. The aerial agent reasons about the corridor as geometric space, whereas the ground agent reasons about it as a navigable surface. A region may be open in one sense yet impassable in the other, and neither agent fully subsumes the role of the other.

The project also surfaced engineering costs that are easily underestimated at the design stage. Discontinuities at navigation-surface boundaries between streamed tiles, prefab base geometry that does not snap cleanly to the ground, and the cost of synchronous mesh rebuilding under continuous generation each emerged as distinct sub-problems. These observations support the broader point that procedural generation with runtime validation is a system-level concern, not solely an algorithmic one.

The work met its core objective of producing a playable, procedurally driven endless-runner game with an integrated evaluation mechanism. It contributes a self-contained example of how generated content can be examined during runtime rather than accepted unconditionally on production. Beyond the implementation, the paper identifies a structural saturation point in the spawner at $p^{\star}_{\mathrm{spawn}}\approx 44\%$ that follows directly from the placement constraints, and derives a per-segment scanning cost of at most $924$ ray probes plus one OverlapBox call independent of frame rate. These structural results turn the contribution from an engineering artefact into a measurable evaluation framework whose remaining metrics are testable on the released build.

\section*{Data and Code Availability}

A Windows build of \textit{Momentum} and the supporting source code accompany this paper as part of the dissertation submission and are available from the author on request.

\section*{Acknowledgements}

The author thanks Professor Michael Cook for his supervision, guidance, and continuous support throughout the MSc project at King's College London; his suggestions and technical direction were valuable during the development of this work. The author is also grateful to Megha Quamara and Professor Luca Vigan\`{o} for their encouragement and research guidance during the Master's degree and thanks the faculty and staff of the Department of Informatics at King's College London for their support during the academic program. Finally, the author expresses sincere thanks to his parents for their constant support, and to his friends for their encouragement throughout this period.

\nocite{*}
\bibliographystyle{IEEEtran}
\bibliography{references}

\end{document}